# Video Anomaly Detection and Localization via Gaussian Mixture Fully Convolutional Variational Autoencoder

Yaxiang Fan, Gongjian Wen, Deren Li, Shaohua Qiu and Martin D. Levine

*Abstract*—We present a novel end-to-end partially supervised deep learning approach for *video anomaly detection and localization* using only normal samples. The insight that motivates this study is that the normal samples can be associated with at least one Gaussian component of a Gaussian Mixture Model (GMM), while anomalies either do not belong to any Gaussian component. The method is based on Gaussian Mixture Variational Autoencoder, which can learn feature representations of the normal samples as a Gaussian Mixture Model trained using deep learning. A Fully Convolutional Network (FCN) that does not contain a fully-connected layer is employed for the encoder-decoder structure to preserve relative spatial coordinates between the input image and the output feature map. Based on the joint probabilities of each of the Gaussian mixture components, we introduce a sample energy based method to score the anomaly of image test patches. A two-stream network framework is employed to combine the appearance and motion anomalies, using RGB frames for the former and dynamic flow images, for the latter. We test our approach on two popular benchmarks (UCSD Dataset and Avenue Dataset). The experimental results verify the superiority of our method compared to the state of the arts.

## I. Introduction

Intelligent video surveillance using computer vision technology to analyze and understand long video streams, plays an irreplaceable role in public security. As an important component of intelligent video surveillance, anomalous event detection automatically discovers and identifies anomalies while monitoring an ever-changing scene and then takes timely measures to deal with emergencies. The major challenge to achieving this is that anomalous events are inherently difficult to define. After all, an anomaly refers to something that is different from the norm. But how different? Our approach to dealing with this issue is to invoke partially supervised learning, which requires only normal samples for training a Deep Neural Network. As a consequence, the samples that are not consistent with the normal samples are considered as anomalies.

In the past, researchers have conducted extensive research in developing so-called "hand-crafted" features to efficiently represent video events. Object trajectories of normal events [28][31] were extracted by employing conventional visual tracking methods to represent the movement of an object. Then, those objects corresponding to trajectories that deviated from the learnt trajectories were considered as anomalies. Since trajectory-based methods were generally found to be impractical for analyzing complex scenes, these methods were replaced by the use of local cuboids to model the trajectory path. These included low-level features such as spatio-temporal gradients [32] [37], histograms of optical flow (HOF) [35], mixture of dynamic textures (MDTs) [34] and acceleration features [9] are extracted from 2-D image patches or local 3-D video blocks.

A significant limitation of the methods based on handcrafted features is that they are difficult to adapt to the huge variations of anomalous events found in different scenes. Recently, following the impressive results of deep architectures on computer vision tasks such as object recognition [12][23], object detection [13] and action recognition [14], attempts have been made to train deep networks for the task of anomalous event detection in video. Motivated by the success of deep learning technology, researchers [15]-[17], [38], [39] began to apply it to anomalous event detection. Most of these methods utilize the deep network as the features extraction and then train detection model, for example, a one-class SVM. However, these deep features are suboptimal because they are not designed or optimized for the whole problem.

Different from these methods mentioned above, we propose an end-to-end deep learning framework for training exclusively on the normal samples. The key idea behind our method is that the normal samples can be associated with at least one Gaussian component of the Gaussian Mixture Model (GMM). Then a test sample that can not be associated with any Gaussian component is identified as anomaly. Our method is based on the Gaussian

Y. Fan, G. Wen and S. Qiu are with the Science and Technology on Automatic Target Recognition Laboratory (ATR), National University of Defense Technology, Changsha, Hunan, 410073, China. E-mail: {(fanyaxiang@126.com), (wengongjian@sina.com), (qiush125@163.com)}

D. Lee is with State Key Laboratory of Information Engineering in Surveying, Mapping and Remote Sensing, Wuhan University, Wuhan, Hubei, 430071, China. E-mail: (drli@whu.edu.cn)

M. D. Levine is with the Department of Electrical and Computer Engineering, Center for Intelligent Machines, McGill University, 3480 University Street, Montreal, QC H3A 2A7, Canada. E-mail: (levine@cim.mcgill.ca)

Mixture Variational Autoencoder[1] [7], which is a model for probabilistic clustering within the framework of Variational Autoencoder (VAE) [2]. Similar as the Autoencoder [1], it contains the encoder-decoder structure that permits learning a mapping from high dimensional data to a low-dimensional latent representation while ensuring a high reconstruction accuracy. Furthermore, the low-dimensional latent representation is constrained to be a Gaussian Mixture Model (GMM). The encoder-decoder structure and Gaussian Mixture constraint of the latent representations correspond to two main components of anomaly detection [8]: feature extraction and model construction. In fact, these two components are joint optimized in our method, which can maximize the performance of the joint collaboration. A fully Convolutional Network (FCN) that does not contain a fully-connected layer is employed for the encoder-decoder structure to preserve relative spatial coordinates between the input image and the output feature map. Over all, we called the deep network, a Gaussian Mixture Fully Convolutional Variational Autoencoder (GMFC-VAE).

Inspired by the human vision system and the two-stream hypothesis [41], we employ a two-stream framework that has already yielded satisfactory results on video recognition tasks such as action recognition [14], action detection [40] and anomalous event detection [15]. In detail, we use the GMFC-VAE to computationally simulate the two data pathways. The spatial stream operates on RGB frames and captures the appearance anomalies. For the temporal stream, dynamic flows[2] [20], that is generated using a Ranking SVM formulation, instead of the conventional optical flow to capture the motion anomalies. The dynamic flow is an amalgamation of a number of *sequential optical flow* frames and can capture long-term temporal information, which optical flow cannot do.

In general, our proposed method includes three stages: training, testing and integrating. In the training stage, *image patches* of both RGB images and dynamic flows are densely sampled, and used as input for the two separate GMFC-VAE networks. This provides an opportunity to simultaneously learn both the latent representation and the Gaussian Mixture Model of the latent representation. Then during the testing stage, the latent representations of the RGB frame patches and the dynamic flow patches are obtained from the two GMFC-VAEs. This permits the computation of the conditional probability of the test patches that belong to each of the components of the Gaussian mixture model. A sample energy based method is used to detect both the appearance and motion anomalies by invoking the joint probabilities. Accordingly, all of the anomalous events are located based on both object motion and appearance. We conduct experiments on two widely available public datasets. The results of the experiments indicate that our method is very competitive compared to state-of-the-art algorithms.

In summary, the main contributions of our work are as

follows:

- The detection of anomalies in a video is based on the hypothesis that the normal samples can be associated with at least one Gaussian component of the Gaussian Mixture model (GMM), while a test sample which is not associated with Gaussian components is declared to be anomaly. This is achieved by an end-to-end deep learning framework, which we refer to as a Gaussian Mixture Fully Convolutional Variational Autoencoder (GMFC-VAE). The latter is established based on the normal samples, learning feature representations of the normal samples as a Gaussian Mixture Model. To the extent of our knowledge, this is the first time that a Variational Autoencoder (VAE) framework has been considered for video anomaly detection.

- Instead of the usual optical flow, we adopted popular two-stream network to employ dynamic flows for detecting the motion anomalies.

- A sample energy based method is proposed to detect anomalies based on the joint probabilities of all of the components in the Gaussian Mixture Model.

- Experiments are used to evaluate our approach on two public datasets. These demonstrate the superiority of our method compared to the state of-the-art methods.

The remainder of this paper is organized as follows. Section II reviews the related literature. Section III is a detailed presentation of the proposed approach. Experimental evaluation is given in Section IV. Finally, Section V concludes the paper.

## II. RELATED WORK

Anomalous event detection and localization has been extensively studied in computer vision for the past 10 years and a wide variety of methods has been introduced. There exist two main categories of anomalous event detection and localization methods: methods based on handcrafted features and deep learning. These are reviewed in this section. Then we present a brief introduction of the Variational Autoencoder (VAE), the motivation of our work.

### A. Methods Based on Handcrafted Features

Handcrafted features have been conventionally used to represent an event. According to [27], these methods can be divided into two categories: those based on trajectories and those on cuboids.

For the trajectory methods, object trajectories are first extracted using object detection and tracking. Each trajectory represents the movement of an object as a sequence of image coordinates. Thus the assumption is made that anomalous trajectories differ from normal ones. For example in [28], features are extracted from the object trajectories in traffic sequences and then clustered by a Mean-Shift Algorithm.

---

[1] The approach is called Variational Deep Embedding (VaDE) in [7]. However, by consulting GMVAE [45] and Rui Shu's blog [46], which involved work similar to [7], we have chosen to call it the Gaussian Mixture Variational Autoencoder for easy understanding.

[2] To be distinguished from the terms optical flow and dynamic image.

Those trajectories that are far from cluster centers in feature–space are defined as being an anomaly. Similarly, Ouivirach et al. [29] have proposed a method for tracking foreground blobs to obtain trajectories and then automatically train a bank of linear HMMs based on the trajectories. An anomalous event is then identified by an analysis of the patterns generated by these scene-specific statistical models. In [30], the anomalies are defined at multiple semantic levels, such as point anomaly of a video object, sequential anomaly of an object trajectory, and co-occurrence anomaly of multiple video objects. Then frequency-based analysis is exploited to automatically discover regular rules for normal events. Test samples that violate these rules are identified as anomalies. Recently, Bera [31] proposed an algorithm based on trajectory-level behavior learning. The anomaly is determined by measuring the Euclidean distance between the local and global pedestrian features of each pedestrian.

Trajectory-based methods have achieved satisfactory performance for anomalies based on speed and direction. However, most of these have relied on object detection and tracking procedures, which are generally not very robust for crowded scenarios. Local cuboid-based features have also been proposed. Instead of object trajectories for representing events, these employ local features such as histograms of gradients (HOG), histograms of optical flow (HOF), spatio-temporal gradients extracted from local 2-D image patches or local 3-D video cuboids. In [30], the variations in local spatial-temporal gradients are used to represent the video events; anomalous events are detected using distribution-based hidden and coupled hidden Markov models. Based on the interaction forces of individuals in a group, Mehran et al. [33] have suggested representing an event by a social force model (SFM) to capture the dynamics of crowd behavior; a bag of words approach is employed to distinguish anomalous frames from the normal ones. Mahadevan et al. [34] proposed the use of a set of mixture of dynamic textures models to jointly model the dynamics and appearance in crowded scenes. In [35] and [36], the Multi-scale Histogram of Optical Flow (MHOF) is extracted to represent an event and anomalies are detected based on a sparse reconstruction cost (SRC). Lu et al. [37] updated the SRC detection model to contain a sparse combination of learning and 3D gradient features to represent an event. Similarly, Giorno et al. [44] proposed detecting changes in video clips by finding frames that can be distinguished from previous frames. As an extension of the Bag of Video words (BOV) approach, Roshtkhari and Levine [42] introduced a probability density function to encode spatio-temporal configurations of video volumes based on spatio-temporal gradient features. By combining statistical feature such as HOG and HOF together to represent the events, Yuan et al. [27] detected anomalous events based on a statistical hypothesis test. By including velocity and entropy information to HOF, Colque et al. [43] proposed a new spatiotemporal feature descriptor, called Histograms of Optical Flow Orientation and Magnitude and Entropy (HOFME). In general however, although local cuboid-based methods are robust when dealing with complex scenes, one disadvantage of these methods is that it may fail to detect the long term activities such as loitering. That is because loitering is related to global movement of a person in the long term rather than the very local cuboid movements.

### B. Deep Learning Methods

Deep learning has recently been used for anomaly detection and has produced state of the art results. The first work that applied deep learning to anomaly detection was [16], which was based on a cascade of auto-encoders. It employed the reconstruction error of the auto-encoder as well as a sparseness measurement of a sparse auto-encoder. Ravanbakhsh et al. [38] used a Fully Convolutional Network as a pre-trained model and inserted an effective binary quantization layer as the final layer of the net to capture temporal CNN patterns. By combining these temporal CNN patterns with a hand-crafted feature (optical flow), they proposed a new measure for detecting local anomalies. Sabokrou et al. [39] employed fully convolutional neural networks (FCNs) to extract discriminative features of video regions. They modeled a normal event as a Gaussian distribution and labeled a test region that differed from the normal reference model as anomaly. Recently, Hasan et al. [17] employed both a fully-connected auto-encoder and a fully-convolutional auto-encoder to learn temporal regularity. However, decisions were based on handcrafted features and short video clips, respectively. A regularity score was computed from the reconstruction errors to detect the anomalies.

The most similar to our paper is the work by Xu et al. [15], which proposed a three-stream architecture (spatial, temporal and their joint representation) by employing the auto-encoder to learn the features. Following this, a one-class support vector machine was exploited to predict the anomaly scores for each stream. A late fusion strategy was then applied to integrate the three-stream scores to make a final decision. The primary difference from our approach is that we do not need to train one-class SVMs or any other event detection model in addition to the learned visual representations. In fact, our approach is an end-to-end deep learning framework, which learns the feature representations of the normal samples as a Gaussian Mixture Model (GMM) by using deep learning. Moreover, we employ so-called *dynamic flow images* instead of the usual optical flow images to represent the motion information.[3]

### C. Variational Autoencoder

An Autoencoder [1] learns a **latent representation z** for a set of data **x** by aligning the outputs **x̄** of the Autoencoder to be equal to the inputs **x** . An Autoencoder consists of an encoder and a decoder.

In addition, by assuming that the latent representation **z** accords with a Gaussian distribution, a Variational Autoencoder (VAE) [2] produces a generative model that creates something very similar to the training data. By inheriting the architecture of a traditional Autoencoder, a Variational Autoencoder consists of two neural networks:

1) Recognition network (encoder network)*: a probabilistic*

---

[3] See Section III.A for a discussion of *dynamic flow images*.

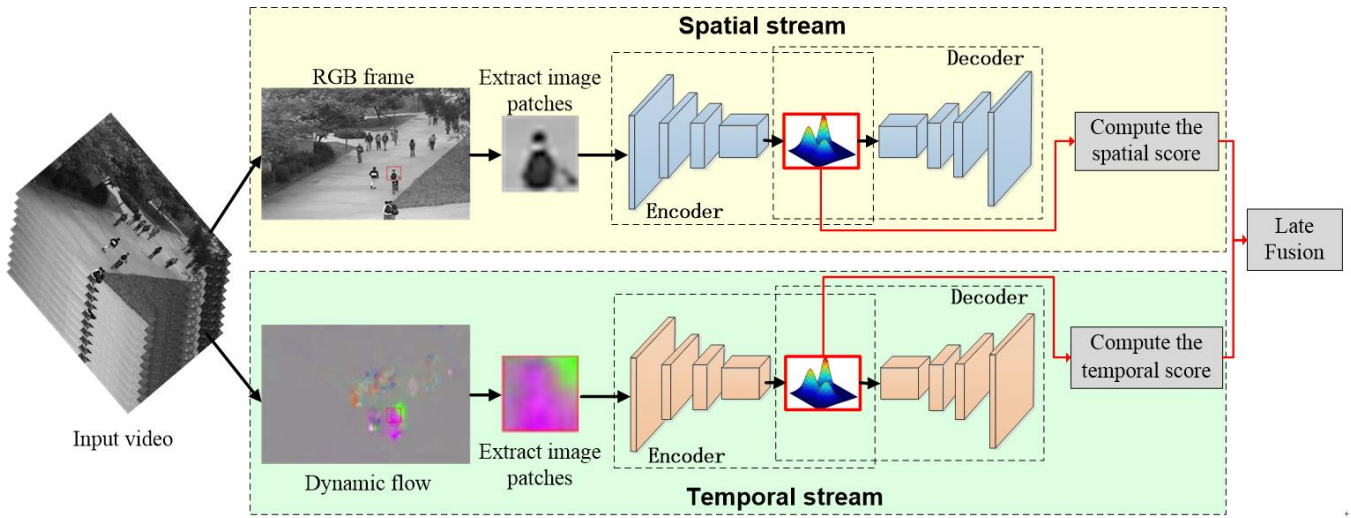

Fig. 1. Overview of the proposed method for anomalous event detection.

encoder $g(\bullet;\phi)$, which map input $\mathbf{x}$ to the latent representation $\mathbf{z}$ to approximate the true (but intractable) posterior distribution $p(\mathbf{z}|\mathbf{x})$,

$$\mathbf{z}=g(\mathbf{x};\phi) \tag{1}$$

2) Generative network (decoder network): *a generative decoder* $f(\bullet;\theta)$, which reconstructs the latent representation $\mathbf{z}$ to the input value $\tilde{\mathbf{x}}$ and does not rely on any particular input $\mathbf{x}$,

$$\tilde{\mathbf{x}}=f(\mathbf{z};\theta) \tag{2}$$

where $\phi$ and $\theta$ denote the parameters of these two networks.

The recognition network $g(\bullet;\phi)$ and generative network $f(\bullet;\theta)$ could be represented as $q_\phi(\mathbf{z}|\mathbf{x})$ and $p_\theta(\mathbf{x}|\mathbf{z})$, respectively. According to the variational inference theory [3], the loss function of the Variational Autoencoder is represented as:

$$\mathcal{L}(\theta,\phi,\mathbf{x})=E_{\mathbf{z}\sim q_\phi(\mathbf{z}|\mathbf{x})}[\log p_\theta(\mathbf{x}|\mathbf{z})]-D_{KL}(q_\phi(\mathbf{z}|\mathbf{x})\parallel p(\mathbf{z})) \tag{3}$$

The first term of (3) is the expected log-likelihood of the input $\mathbf{x}$, which encourages the decoder $p_\theta(\mathbf{x}|\mathbf{z})$ to reconstruct the input $\mathbf{x}$. It could be considered as the reconstruction loss and incurs a large value for good reconstructions. The second term of (3) is the Kullback-Leibler divergence [4] between the $q_\phi(\mathbf{z}|\mathbf{x})$ and $p(\mathbf{z})$, which are the distribution, we wish to learn with the encoder and the prior distribution of the latent representation $\mathbf{z}$, respectively. It measures the difference between two probability distributions and produces a small value when their similarity is very strong. Using the so-called "reparameterization trick" [3], the parameters $\phi$ and $\theta$ can be obtained by optimizing (3) via stochastic gradient variational bayes [2].

Not only does that a VAE have the ability to generate a variety of complex data [2][5][6], it has also been shown to be effective for anomaly detection [18][19]. This is based on the assumption that the latent representation of normal samples is consistent with a Gaussian distribution. This implies that all training data samples are clustered in feature space and the anomalies are far from this cluster center. In fact, this hypothesis is not rigorous since the normal samples may indeed cluster around more than one centroid. In order to deal with this issue, we define a Gaussian Mixture Fully Convolutional Variational Autoencoder (GMFC-VAE), which is employed to detect anomalies. We assume that the latent representation of the training samples accords with a Mixture-of-Gaussians Model instead of a simple Gaussian distribution.

## III. METHOD

In this section, we present the proposed approach for anomaly detection and localization in detail. Firstly, based on the Ranking SVM formulation, dynamic flows are generated to represent the motion cue. Then, a two-stream Gaussian Mixture Fully Convolutional Variational Autoencoder (GMFC-VAE) is used to learn an anomaly detection model utilizing the normal samples of RGB images and dynamic flows, respectively. Given a test sample corresponding to an image patch and the two learned models, both the appearance anomaly and motion anomaly scores can be predicted by exploiting a sample energy-based method. Finally, these two complementary cues are fused to achieve the final detection results. The overview of the proposed method is illustrated in Fig. 1.

### A. Obtaining Dynamic Flow

Dynamic flow is an *amalgamation* of a number of sequential frames, based on the optical flow computed for each frame in a video. Compared with the more familiar raw optical flow, which contains only the motion cues between two consecutive frames, dynamic flow is capable of capturing long-term temporal information.

Given a video of length $n$ frames, each of which contain the conventional optical flows $\mathcal{F} = \{f_i\}_{i=1}^n$, where $f_i \in \mathbb{R}^{m_1 \times m_2 \times 2}$, and $m_1$, $m_2$ are the height and width of the image. According to [20], the *horizontal* dynamic flow channel $F^u \in \mathbb{R}^{d_1 \times d_2}$ as well as the *vertical* flow channel $F^v \in \mathbb{R}^{d_1 \times d_2}$ can be approximated by minimizing the upper bound of $\sum \xi_{ij}$, by solving the following problem:

$$\text{Minimize: } L(F^u, F^v, \xi_{ij}) = \|F^u\|^2 + \|F^v\|^2 + C\sum_{i<j} \xi_{ij}$$

Subject to: $\forall i < j$

$$\langle F^u, \overline{f_i^u} \rangle + \langle F^v, \overline{f_i^v} \rangle \le \langle F^u, \overline{f_j^u} \rangle + \langle F^v, \overline{f_j^v} \rangle + 1 - \xi_{ij} \quad (4)$$

where $\xi_{ij}$ is a slack variable and $\forall(i,j): \xi_{ij} \ge 0$. $C$ is a soft margin parameter that controls the trade-off between margin size and training error. $f_i^u$ and $f_i^v$ represent the horizontal and vertical components of the optical flow image $f_i$, respectively. Given that $\overline{\cdot}$ represents an averaging operation, then the Time-Varying Means (TVM) embody the flow images. $\overline{f_i^u} = \sum_{t=1}^i f_t^u$, $\overline{f_j^u} = \sum_t^j f_t^u$, $\overline{f_i^v} = \sum_t^i f_t^v$, and $\overline{f_j^v} = \sum_t^j f_t^v$. We also note that $\langle \bullet, \bullet \rangle$ signifies the inner product of the TVM flow image and the dynamic flow that are to be found.

Equation (4) can be solved by training a linear ranking machine, such as RankSVM [21]. We observe that this will facilitates the conversion of a number of sequential optical flow frames to a two channel dynamic flow. For each optical flow frame $f_i$, we compute the dynamic flow $F_i$ from the sequence of $\{f_{i'}\}_{i'=i}^{i+\Delta t}$, where $\Delta t$ is the window size. Thus, a video of length $n$ optical flow frames, $\mathcal{F} = \{f_i\}_{i=1}^n$, can be converted to a set of dynamic flows $\{F_i\}_{i=1}^{n-\Delta t}$ to represent the motion in the whole video. Thus the motion can represented by the dynamic flow $F_i$ instead of commonly used optical flow $f_i$.

It can be observed in Fig. 2 that the dynamic flow images indicate an unbelievable ability to characterize events. The motion cues are represented by two regions of similar shape but different colors that indicate the beginning and evolution of the motion. In addition, the intensity of the color denotes the degree of movement. Compared with optical flow, which captures motion cues between two consecutive frames, the objects in the dynamic flow images are more salient and represent long-term temporal information.

### B. Learning Appearance and Motion Anomaly Detection Models Using Gaussian Mixture Fully Convolutional Variational Autoencoders

In this section, we present how to learn the appearance and motion anomalous detection models with Gaussian Mixture Fully Convolutional Variational Autoencoders (GMFC-VAE). Following the same strategy as [15], we exploit the appearance cue (RGB frames) and motion cue (dynamic flows) to detect anomalies in both these domains.

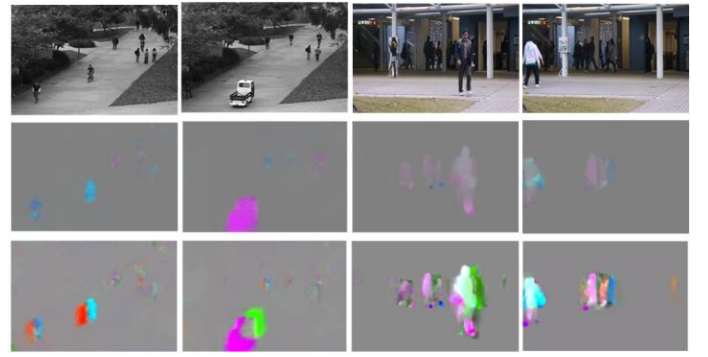

Fig. 2. Examples of dynamic flow image, optical flow and related RGB video frames. The first row shows a single frame from four different videos. The second row indicates the optical flow for the frames in the first row. Finally, the third row presents the dynamic flow for each of the complete videos. Thus the dynamic flow "summarizes" the overall behavior of all of the moving objects in the scene. Also note that the background has been removed.

We train two separate models for RGB and dynamic flow as inputs. A set of training patches $\mathbf{x}' = \{x_i\}_{i=1}^n$ (RGB image patches or dynamic flow patches) are obtained by a sliding window from the training set of videos, where $n$ is the number of the training patches. The size of each patch $x_i$ is $weith \times height \times channel$, where $weith$, $height$, and $channel$ represent the width, height and channel, respectively, of the patch. All patches are linearly normalized into a range of $[0,1]$ and employed as the input for training the GMFC-VAE.

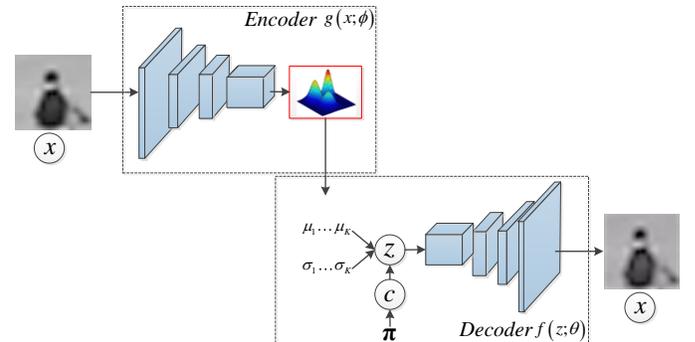

Fig. 3. The diagram of GMFC-VAE.

Variational autoencoder typically assume that the priors of the latent representation $\mathbf{z}$ follow a simple Gaussian distribution. However, here we assume that a Mixture-of-Gaussians (MoG) is the prior of the latent representation [7]. Consequently, the GM-VAE can be applied to describe the distribution of the normal samples. For normal samples $\mathbf{x}'$ with latent representation $\mathbf{z}'$, as shown in Fig. 3, the generative model can be reformulated as three steps: 1) choose a Gaussian mixture $c$; 2) obtain a latent vector $\mathbf{z}'$; 3) according to the latent representation $\mathbf{z}'$ obtain the reconstruction result $\mathbf{x}'$. It can be denoted as follow:

$$c \sim \text{Category}(\boldsymbol{\pi}) \tag{5}$$

$$\mathbf{z}' \sim \mathcal{N}(\mu_c, \sigma_c^2\mathbf{I}) \tag{6}$$

$$p(\mathbf{z}', c) = \pi_c \mathcal{N}(\mathbf{z}' \mid \mu_c, \sigma_c^2) \tag{7}$$

$$[\mu_x; \log \sigma_x^2] = f(\mathbf{z}'; \theta') \tag{8}$$

$$\mathbf{x}' \sim \mathcal{N}(\mu_x, \sigma_x^2 I) \tag{9}$$

where $K$ is a predefined number of components of the mixture and $\boldsymbol{\pi} = [\pi_1, \pi_2, \cdots, \pi_K]$ is the prior probability of the Gaussian mixture components, such that $\pi_1 + \pi_2 + \cdots + \pi_K = 1$. The value of $K$ is discussed in Section IV-E. The $c^{th}$ vector component is characterized by normal distributions with means $\mu_c$ and covariance $\sigma_c^2$. $\mathbf{I}$ is an identity matrix. $f(\bullet; \theta')$ is the *decoder* parametrized by $\theta'$, $\mathcal{N}(\mu_x, \sigma_x^2 I)$ is Gaussian distribution parametrized by $\mu_x$ and $\sigma_x^2$.

Similar to VAE, the *encoder* $g(\bullet; \phi')$ is used for approximate true posterior $p(\mathbf{z}, c \mid \mathbf{x})$,

$$[\tilde{\mu}, \log \tilde{\sigma}^2] = g(\mathbf{x}; \phi') \tag{10}$$

And the *encoder* $g(\bullet; \phi')$ and the decoder $f(\bullet; \theta')$ could be represented as $q_{\phi'}(\mathbf{z}' \mid \mathbf{x}')$ and $p_{\theta'}(\mathbf{x}' \mid \mathbf{z}')$, respectively. And from Equation (8)(9)(10), we obtain that

$$q_{\phi'}(\mathbf{z}' \mid \mathbf{x}') = \mathcal{N}(\mathbf{z}' \mid \tilde{\mu}, \tilde{\sigma}^2 I) \tag{11}$$

$$p_{\theta'}(\mathbf{x}' \mid \mathbf{z}') = \mathcal{N}(\mathbf{x}' \mid \mu_x, \sigma_x^2 I) \tag{12}$$

It follows that the loss function of the GM-VAE can be denoted as:

$$\mathcal{L}(\theta', \phi') = E_{\mathbf{z}' \sim q_{\phi'}(\mathbf{z}', c \mid \mathbf{x}')}[\log p_{\theta'}(\mathbf{x}' \mid \mathbf{z}')]$$
$$- D_{KL}(q_{\phi'}(\mathbf{z}', c \mid \mathbf{x}') \parallel p(\mathbf{z}', c)) \tag{13}$$

where $\phi'$ and $\theta'$ are the parameters of the encoder and decoder, respectively. The first term in (13) can be regarded as a reconstruction cost. The second term is given by the Kullback-Leibler divergence between the Mixture-of-Gaussians (MoG) prior $p(\mathbf{z}', c)$ and the variational posterior $q(\mathbf{z}', c \mid \mathbf{x}')$.

By substituting the terms in Equation (13) with Equations (5), (7), (11) and (12), and using the SGVB estimator, the loss function can be written as:

$$\mathcal{L}(\theta', \phi', \pi, \mu_c, \sigma_c) = \frac{1}{L} \sum_{i=1}^{L} \|x_i - \tilde{x}_i\|_2^2$$
$$- \frac{1}{2} \sum_{c=1}^{K} \gamma_c \left( \log \sigma_c^2 + \frac{\tilde{\sigma}^2}{\sigma_c^2} + \frac{(\tilde{\mu} - \mu_c)^2}{\sigma_c^2} \right)$$
$$+ \sum_{c=1}^{K} \gamma_c \log \frac{\pi_c}{\gamma_c} + \frac{1}{2}(1 + \log \tilde{\sigma}^2) \tag{14}$$

where $L$ is the number of Monte Carlo samples in the SGVB estimator, $x_i$ is the $i$-th training patch, $\tilde{x}_i$ is the construct result of $x_i$, $K$ is a predefined number of components of the mixture, $\pi_c$ is the prior probability of the $c$-th Gaussian mixture components, and $\gamma_c$ denotes for $q(c \mid \mathbf{x}')$ for simplicity. And $q(c \mid \mathbf{x}')$ could be computed as follow:

$$q(c \mid \mathbf{x}') = p(c \mid \mathbf{z}') = \frac{p(c) p(\mathbf{z}' \mid c)}{\sum_{c'=1}^{K} p(c') p(\mathbf{z}' \mid c')} \tag{15}$$

The details for optimizing of training stage can be found in [7].

It has been observed [22] that the use of a convolution layer to replace both the max pooling and fully connected layers of standard CNNs outperforms the state of the art on several object recognition datasets. That is because both a fully connected layer as well as a max pooling layer cause a loss of spatial information while a convolution layer is able to maintain the spatial information. Accordingly, we employ a fully convolutional model for the encoder-decoder network. We refer to this network as the as Gaussian Mixture Fully Convolutional Variational Autoencoder (GMFC-VAE). The details of the network architecture are found in Section IV-C.

### C. Prediction

In last sub-section, we can obtained all the parameters ($\{\theta', \phi', \pi, \mu_c, \sigma_c\}$, $c \in \{1, \ldots, K\}$) of GMFC-VAE. In this sub-section, we discuss how to detect an anomaly after having trained a GMFC-VAE of both the spatial and temporal streams. For the testing phase with test image patch $\mathbf{y}$ (RGB image patch or dynamic flow patch), the latent representation $\mathbf{z}'$ can be achieved by the encoder $q_{\phi'}(\mathbf{z}' \mid \mathbf{y})$. The $p(\mathbf{z}' \mid c)$ can be computed according to (6) as

$$p(\mathbf{z}' \mid c) = \frac{1}{\sqrt{2\pi}\sigma_c} \cdot e^{-\frac{(\mathbf{z}' - \mu_c)^2}{2\sigma_c^2}} \tag{16}$$

If query sample $\mathbf{y}$ is normal, its latent representation $\mathbf{z}'$ must be associated with at least one Gaussian component $i$ ($i = 1, 2, 3, \ldots, K$) of the training data, which would produce a relatively high conditional probability $p(\mathbf{z}' \mid i)$. On the other hand, the conditional probability for the other Gaussian components $j$ ($j = 1, 2, 3, \ldots, K, j \neq i$) would be relatively low.

In contrast, the latent representation of an anomalous query sample would most likely not be associated with any of Gaussian components. This would also engender a low conditional probability $p(\mathbf{z}' \mid c')$ for all of the Gaussian components $c'$ ($c'$=1,2,3,..., $K$).

Motivated by [47][48], the anomaly score of the query test sample is computed by an sample energy method in form of the log-likelihood:

$$
\begin{aligned}
E(\mathbf{z}') &= -\log\left(\sum_{c'=1}^{K} p(\mathbf{z}' \mid c') p(c')\right) \\
&= -\log\left(\sum_{c'=1}^{K} \pi_{c'} \cdot \frac{1}{\sqrt{2\pi}\sigma_{c'}} \cdot e^{-\frac{(\mathbf{x}' - \mu_{c'})^2}{2\sigma_{c'}^2}}\right)
\end{aligned}
\tag{17}
$$

It is obvious that the anomalies have the higher scores. Suppose the appearance and motion anomaly scores are labelled $\mathcal{S}_{appearance}$ and $\mathcal{S}_{\text{motion}}$. Then the overall anomaly score, $\mathcal{S}_{overall}$, is their combination with the importance factors, $\alpha$ and $\beta$:

$$
E_{overall} = \alpha E_{appearance} + \beta E_{\text{motion}}
\tag{18}
$$

Finally, we identify $\mathbf{y}$ is as an anomaly if the following criterion is satisfied:

$$
E_{overall} > \theta
\tag{19}
$$

where $\theta$ is a threshold that determines the sensitivity of the anomalous detection method. A discussion of these tests are found in Section IV-E.

## IV. Experiments

To evaluate both the qualitative and quantitative effectiveness of the proposed algorithm, comparisons with state-of-the-art algorithms, we perform experiments with two public datasets, the UCSD and Avenue Datasets. In this section, we present the datasets, evaluation criteria, details of the experimental settings and experimental results.

### A. Datasets

The UCSD dataset contains two subsets, Ped1 and Ped2, which were recorded at two different scenes by a fixed camera. In detail, the Ped1 Dataset contains 34 normal and 36 abnormal video clips of $238 \times 158$ pixels and each of the video clips contains 200 frames. As for the Ped2 Dataset, it consists of 16 normal and 14 abnormal video clips of size $320 \times 240$ pixels. The length of each video clip in the UCSD Ped2 Dataset is between 150 to 200 frames. For both the Ped1 and Ped2, the normal events contain pedestrians on the walkways, while the abnormal events include bikes, skaters, small cars, and people walking across a walkway or in the grass that surrounds the Walkway. For the two subsets, frame-level ground-truth is provided in the form of a binary flag per frame. In addition, 10 test clips from Ped1 and 12 from Ped2 are provided with pixel-level ground-truth.

The Avenue Dataset contains 15 normal and 21 abnormal videos clips of size $640 \times 360$ pixels, which were recorded in front of school corridors using a fixed camera. Each video clip is approximately 1 to 2 minutes long (25 frames/second). Object-level ground-truth (labeling anomalies with rectangular regions) is provided for this dataset. The normal events contain pedestrians walking in parallel to the camera plane, while the anomalous events contain people running, throwing objects and loitering.

### B. Evaluation Criteria

To compare with existing methods for anomaly detection, we used two evaluation criteria, Frame-Level and Pixel-Level, which currently are widely used in anomaly detection research. The details of the two evaluation criteria are as follow:

1) *Frame-level criterion*: A detected anomalous frame is true positive if it contains at least one anomalous pixel.

2) *The Pixel-level criterion:* A detected anomalous frame is true positive if there is more than 40% overlap with a ground truth region is detected as an anomaly region. This criterion can be used to evaluate the anomaly localization capability.

The *Receiver Operating Characteristic (ROC)* curve of *True positive rate (TPR)* versus *False positive rate (FPR)* is used to measure the accuracy [34], where *TPR* represents the rate of correctly labeled and *FPR* represents the rate of incorrectly labeled frames.

Two evaluation criteria are select as quantitative indexes based on the ROC curves:

1) *Area Under Curve (AUC)*: Area under the ROC curve.

2) *Equal Error Rate (EER)*: The ratio of misclassified frames when the *FPR* equals the miss rate, i.e., the *FPR* at which $FPR = 1 - TPR$.

### C. Implementation Details

1) *Experimental Setup*

First, all of the frames are resized to $420 \times 280$. To construct the dynamic flow, we first compute optical flow for each consecutive pair of frames, according to [24]. Following [18], the values of $f_i^u$ and $f_i^v$ are transformed into the discrete range [0, 255] by employing $f_i^u = f_i^u \times a + b$ and $f_i^v = f_i^v \times a + b$, with $a = 16$ and $b = 128$. The window size $\Delta t$ that generates the dynamic flows is set to $\Delta t = 20$. Then $f_i^u$, $f_i^v$ are stacked to form a two-channel image and input to (4) to generate a single dynamic flow. Then the flow magnitude $F^m$ is computed as the third channel of the dynamic flow using $F^m = \sqrt{(F^u)^2 + (F^v)^2}$.

To detect the appearance and motion anomalies, two distinct GMFC-VAEs are placed at each input to GMFC-VAEs. Two sets of frames, one containing the RGB and the other, the dynamic flow image data of normal samples are supplied as input. These image frame samples are divided into small patches of size $28 \times 28$ with a stride $d_1 = 7$. We eliminate the massive number of patches that do not contain any moving pixels based on a frame difference. This set is then randomly sampled to provide 960K training patches. In the testing stage, the test patches are generated by use sliding windows with a size of $28 \times 28$ and a stride of $d_2 = 28$. That implies that a test frame outputs a score map of resolution $15 \times 10$, thereby splitting each frame into a grid of 150 square samples. We arbitrarily select 0.5 for both $\alpha$ and $\beta$ in (18).

The Encoder and Decoder are pre-trained utilizing a stacked Auto-Encoder which has the same network architecture as the Encoder and Decoder. The parameters of the network were optimized using the Adam optimizer [25] with a learning rate of 0.0001, a momentum of 0.9, a weight decay of 0.0005, and mini-batches of size 100. The proposed method is implemented in Python and Keras [26], which used a Theano backend. The number of mixture components $K$ was initialized as $K = 20$ and then updated according to the discussion in Section IV-E.

### 2) *Network* architecture

We note that the architecture of the encoder resembles the convolutional stage of Model C in [22]. In detail, the encoder has four convolution layers and the size of the first three convolution kernels is $3 \times 3$. The first convolutional layer has 32 filters with a stride of 2 and generates 32 feature maps with a resolution of $14 \times 14$. The second and third convolutional layers contain 64 and 128 filters, respectively, with a stride of 2. The resolution of the output feature maps of the second and the third layers are $7 \times 7$ and $4 \times 4$, respectively. This is followed up with a fourth convolution layer, which comprises 256 filters of size $4 \times 4$. It generates 256 feature maps with a resolution of $1 \times 1$ and can be converted to a 256-D feature. Each convolutional layer is followed by a ReLU nonlinearity.

The Decoder comprises the reverse architecture of the encoder. Two fully-connected layers are placed in parallel at the end of the encoders and result in the means $\mu_c$ and covariance $\sigma_c$ of each of the $c^{th}$ components, respectively. A new sampling layer follows the two fully-connected layers to compute the latent representation $\mathbf{z}'$, as described in (5) and (6). Finally, the latent representation $\mathbf{z}'$ is input to the decoder to obtain the appropriate reconstructed image patch. The detailed configurations of the whole network architecture are

TABLE I
SPECIFICATIONS OF THE GMFC-VAE MODEL

| layer | input | kernel | stride/ pad | output | last/ next layer |
|---|---|---|---|---|---|
| I0 | $3 \times 28 \times 28$ | N/A | N/A | $3 \times 28 \times 28$ | N/A |
| C1 | $3 \times 28 \times 28$ | $3 \times 3$ | 2/0 | $64 \times 14 \times 14$ | I0/C2 |
| C2 | $64 \times 14 \times 14$ | $3 \times 3$ | 2/0 | $128 \times 7 \times 7$ | C1/C3 |
| C3 | $128 \times 7 \times 7$ | $3 \times 3$ | 2/1 | $256 \times 4 \times 4$ | C2/C4 |
| C4 | $256 \times 4 \times 4$ | $4 \times 4$ | 1/0 | $64 \times 1$ | C3/F5 |
| F5 | $64 \times 1$ | $1 \times 1$ | 1/0 | $64 \times 1$ | C4/S7 |
| F6 | $64 \times 1$ | $1 \times 1$ | 1/0 | $64 \times 1$ | C4/S7 |
| S7 | $64 \times 1$ | N/A | N/A | $64 \times 1$ | F5&F6 /D8 |
| D8 | $64 \times 1$ | $4 \times 4$ | 1/0 | $256 \times 4 \times 4$ | D7/D9 |
| D9 | $256 \times 4 \times 4$ | $3 \times 3$ | 2/1 | $128 \times 7 \times 7$ | D8/D10 |
| D10 | $128 \times 7 \times 7$ | $3 \times 3$ | 2/0 | $64 \times 14 \times 14$ | D9/D11 |
| D11 | $64 \times 14 \times 14$ | $3 \times 3$ | 2/0 | $3 \times 28 \times 28$ | D10/O12 |
| O12 | $3 \times 28 \times 28$ | | | | |

I =input layer, C=convolutional layer, F=fully connected layer, S=sampling layer, D=deconvolutional layer, O=output layer
The Encoder and Decoder consist of I0, C1, C2, C3, C4 and D8, D9, D10, D11, O12, respectively.

shown in Table I.

### D. *Experimental results*

### 1) *UCSD dataset*

Consider the qualitative behavior of the detection performance of the UCSD Ped1 and Ped2 datasets in Figs. 4 and 5. The corresponding ROC curves for pixel- and frame-level behaviors are displayed by varying the threshold parameter $\theta$. The ROC curve for several methods are provided for comparison, including seven methods that use handcrafted features [27][33][35][37][42][43] and four that use deep learning [15][16][17][39]. The results of these contrast methods are obtained in their respective paper. A quantitative comparison in terms of the Area Under the Curve (AUC) and the Equal Error Rate (EER) are shown in Table III. From an examination of Table III, it is quite obvious that the use of deep learning features outperforms employing handcrafted features.

The ROCs of Fig. 4 show that our method is comparable to other methods on the UCSD ped1 dataset. Based on frame-level evaluation, our method achieved 94.9% AUC and 11.3% EER on this dataset. This outperforms all of the methods used for comparison. For the pixel level evaluation, our method achieved 91.4% AUC and 36.3% EER, which is better than the other methods except for Statistical Hypothesis Detector [27]. In detail, the Statistical Hypothesis Detector [27] ahead by 1.7% and 5.8% of AUC and EER to our method. Compared with the Statistical Hypothesis Detector [27], as shown in Fig. 4 (b), our method achieves a relatively higher True Positive Rate (TPR) at a low False Positive Rate (FPR). This is crucial for a practical detection system. More quantitative results for frame-level evaluation and pixel-level evaluation are shown in the 1st, 2nd and 3rd and 4th columns, respectively, of Table III.

The ROCs of the UCSD ped2 dataset are presented in Fig. 5 and indicate that the proposed method nearly reaches the best of the state-of-the-art. The right side of Table III shows the frame- and pixel-level results for the tested methods. Our frame-level EER is 12.6% whereas the best result of 11% is achieved by Deep-Anomaly [39]. As well, the pixel-level EER is 19.2%, which is 4.2% less than the Deep-Anomaly [39] algorithm. However, it should be noted that Deep-Anomaly [39] is a combination of a pre-trained CNN (i.e., AlexNet) and a new convolutional layer. Consequently, it is not trained end-to-end and the generalization ability is weak. Results of AUCs show that our method outperforms all the methods (Deep-Anomaly [39] algorithm doesn't provide the AUCs) with respect to both the frame-level and pixel-level measure.

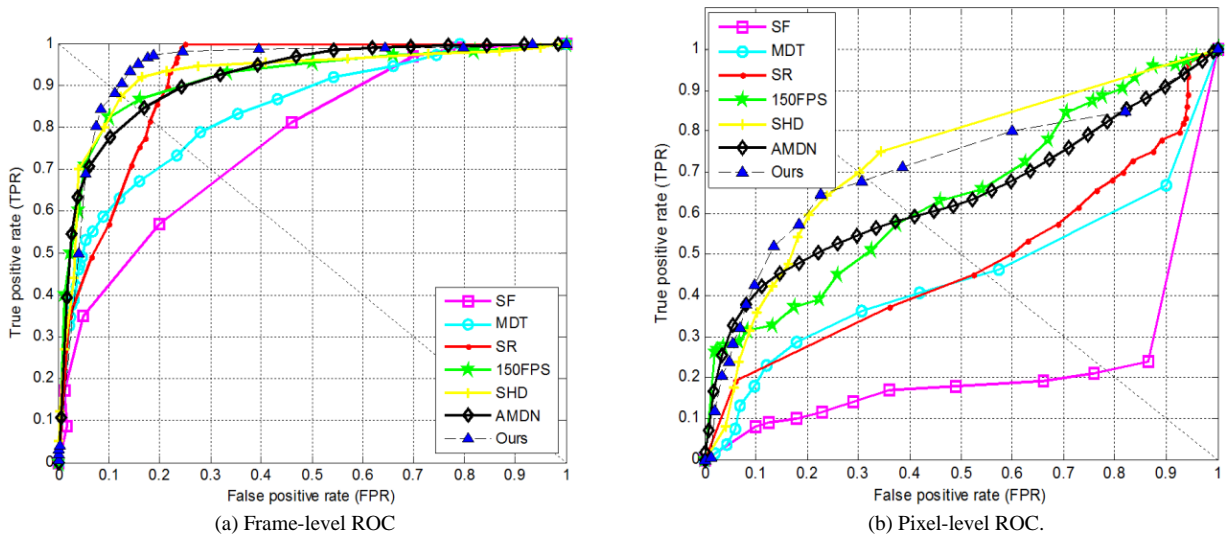

(a) Frame-level ROC      (b) Pixel-level ROC.

Fig. 4. ROC curves for the UCSD Ped1 dataset. Abbreviation: Social Force (SF) [33], MDT [34], Sparse Reconstruction (SR) [35], Detection at 150FPS (150FPS) [37], Statistical Hypothesis Detector (SHD) [27], AMDN [15]

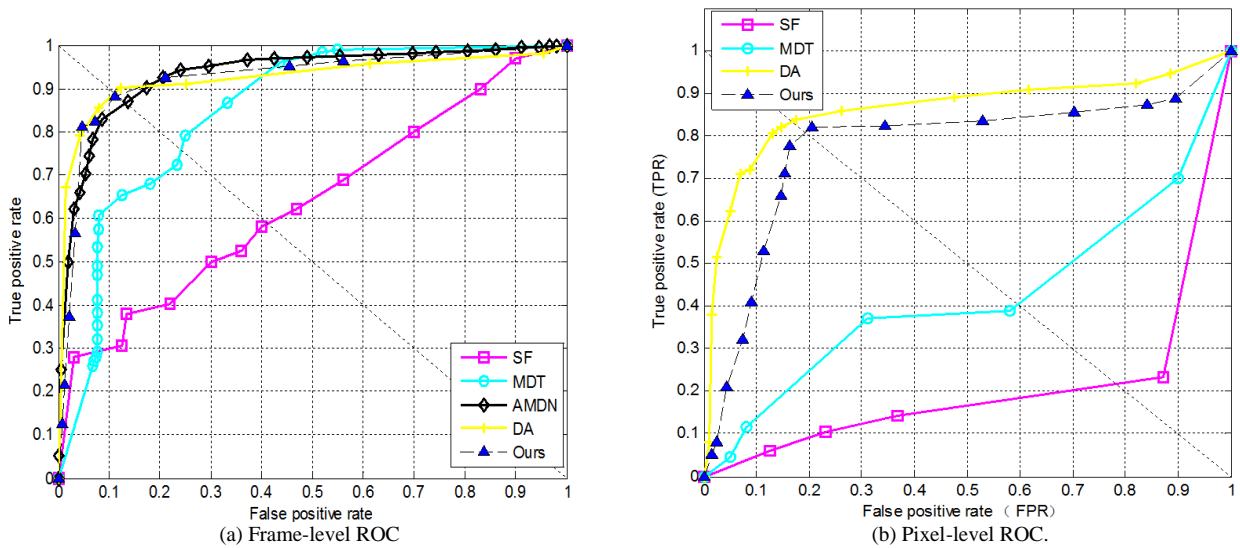

(a) Frame-level ROC      (b) Pixel-level ROC.

Fig. 5. ROC curves for the UCSD Ped2 dataset. Abbreviation: Social Force (SF) [33], MDT [34], AMDN [15], Deep-Anomaly (DA) [39]

TABLE III
COMPARISON WITH THE STATE OF THE ART METHODS
IN TERMS OF AUC% (AREA UNDER ROC) AND EER% (EQUAL ERROR RATE) ON USCD DATASET

| Method | Ped1 (frame level) | | Ped1 (pixel level) | | Ped2 (frame level) | | Ped2 (pixel level) | |
|---|---|---|---|---|---|---|---|---|
| | EER | AUC | EER | AUC | EER | AUC | EER | AUC |
| Social Force (SF) [33] | 31 | 67.5 | 67.5 | 19.7 | 42 | 55.6 | 80 | — |
| MDT [34] | 25 | 81.8 | 58 | 44.1 | 25 | 82.9 | 54 | — |
| Sparse Reconstruction [35] | 19 | — | 54 | 45.3 | — | — | — | — |
| Detection at 150FPS [37] | 15 | 91.8 | 43 | 63.8 | — | — | — | — |
| Dense STV [42] | 16.0 | 89.9 | 57.7 | 41.7 | — | — | — | — |
| Statistical Hypothesis Detector [27] | 12.1 | 93.7 | **30.5** | **73.1** | — | — | — | — |
| HOFME [43] | 33.1 | 72.7 | — | — | 20 | 87.5 | — | — |
| Cascade Auto-encoders [16] | — | — | — | — | 15 | — | — | — |
| Deep-Anomaly [39] | — | — | — | — | **11** | — | **15** | — |
| Learning Temporal Regularity [17] | 27.9 | 81.0 | — | — | 21.7 | 90.0 | — | — |
| AMDN (double fusion) [15] | 16 | 92.1 | 40.1 | 67.2 | 17 | 90.8 | — | — |
| Our Method | **11.3** | **94.9** | 36.3 | 71.4 | 12.6 | **92.2** | 19.2 | **78.2** |

Fig. 6 shows some examples of the detection result on the UCSD dataset, in which detected anomalous events are labeled with red masks. The first row and the second row of Fig. 6 are the results of USCD Ped1 and Ped2, respectively. It is obvious that the proposed method is able to detect different kinds of anomalous events, such as bicycling (Fig. 6 (a) (b) (e) (f) (h)), skateboarding (Fig. 6(e) and Fig. 6(h)), cars (Fig. 6(d) and Fig. 6(g)) and wheelchair (Fig. 6(c)).

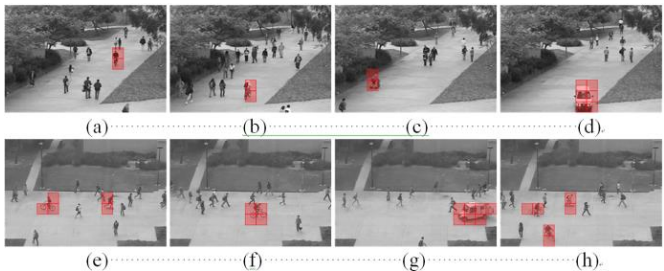

(a)　(b)　(c)　(d)
(e)　(f)　(g)　(h)
Fig. 6. Examples of abnormality detection results on the UCSD dataset

### 2) Avenue dataset

Only a few methods have been tested on the Avenue dataset, which is a new dataset that has been made public recently in [37]. Since the anomalies are labeled by rectangle regions but are not actually rectangular, the ground truth contains background as well as foreground pixels. Because of this, we ignore the Pixel-level measure and use only the Frame-level measure for testing. Three approaches are presented for comparison: they are the Detection at 150FPS [37], Discriminative Framework [44] and Learning Temporal Regularity [17]. The results of these methods are obtained from their respective papers. The Frame-level evaluation, in the form of AUC and EER, are presented in TABLE IV and the ROC curves in Fig. 7.

Compared to the best result of the state-of-art approaches, our method shows an improvement of 2.5% in terms of frame-level AUC. Note that actually only the Learning Temporal Regularity [17] algorithm provides the frame-level EER. However, from the Fig. 7, it can be observed that our method achieves a lower EER than the Detection at 150FPS [35] algorithm and the Discriminative Framework [44] algorithm.

TABLE IV
PERFORMANCE ON THE AVENUE DATASET (AUC% AND EER% AT THE FRAME-LEVEL)

| Method | EER | AUC |
|---|---|---|
| Detection at 150FPS [37] | — | 80.9 |
| Discriminative Framework [44] | — | 78.3 |
| Learning Temporal Regularity [17] | 25.1 | 70.2 |
| Our Method | **22.7** | **83.4** |

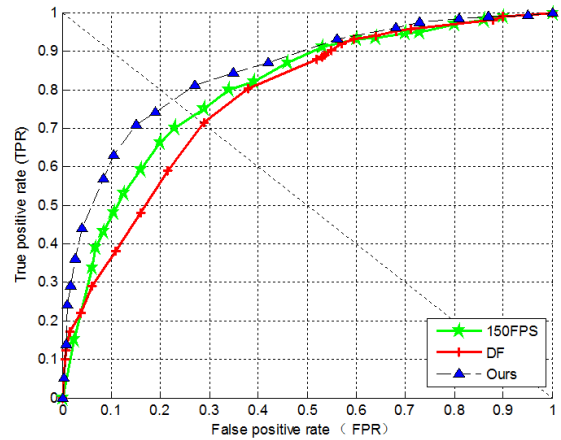

Fig. 7. Frame-level ROC curves for the Avenue Dataset. Abbreviation: Detection at 150FPS (150FPS) [16], Discriminative Framework (DF) [44]

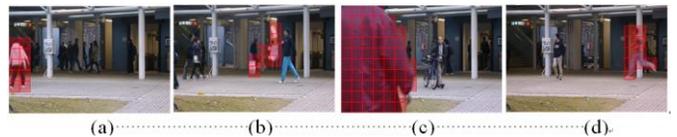

(a)　(b)　(c)　(d)
Fig. 8. Examples of abnormality detection results on the Avenue dataset

Overall, the results prove that our method is very effective on the Avenue dataset. Fig. 8 shows some examples of the detected anomalous events, such as waving hands (Fig. 8 (a)), throwing papers (Fig. 8 (b)), blocking the camera (Fig. 8 (c)) and running (Fig. 8 (d)).

### E. Analysis

#### 1) Evaluation of the number of mixture components

In this sub-section, we analyze the impact of the number of mixture components $K$ [4] on the detection results. Recall that $K$ is pre-defined and decides the number of normal patch clusters. In the experiments, various $K$ values were applied and the Area Under the ROC Curve (AUC) at the frame-level was calculated.

Table V presents the detection performance on the UCSD datasets. We observe that performance on both Ped1 and Ped2 increases with $K$ when it is less than 20. This is due to the fact that small values of $K$ produce less clusters that causes inadvertent clusters of normal patches. Some samples that belong to different categories are grouped into one cluster thereby losing some local information. The performance holds steady with increasing number of mixture components when $K$ is beyond 20. According to (15) and (17), using larger $K$ requires more computational. Therefore $K = 20$ seems to be a reasonable choice by trading-off algorithm performance against computational cost.

#### 2) Evaluation of spatial and temporal streams

To further demonstrate the validity of our method, we evaluate the performance of GMFC-VAE under two different settings in (18): (1) Spatial Stream: Only the appearance cue is used for detection ($\alpha = 1, \beta = 0$); (2) Temporal Stream: ($\alpha = 0, \beta = 1$). We compare the results with our late fusion results ($\alpha = 0.5, \beta = 0.5$) on UCSD ped1 in Table VI. Clearly, the performance of both the Spatial Stream and Temporal Stream are worse than the late fusion result. That is because either the appearance cue or the motion cue is employed in the two cases.

Some examples of the anomalous detection results of the three settings on the UCSD Ped1 dataset are shown in Fig. 9. The anomalous events include: (a) small car, (b) skater, (c) biker and people walking across a walkway or on the grass surrounding it, (d) biker. The four columns display the ground-truth, Spatial Stream detection result, Temporal Stream detection result and late fusion result, respectively.

The Temporal Stream is able to detect the appearance anomalies such as the cars, the skater and the biker (Fig. 9 II-a, II-b, II-d), while missing the skater standing on his skates (Fig. 9 II-b) and the biker on the bike (Fig. 9 II-d). The Temporal Stream also missed the persons walking across a walkway and in the grass surrounding it (Fig. 9 II-c).

For the temporal stream, all the missed patches of the spatial stream are identified, such as the missing skater on his skates (Compare Fig. 9 II-b and Fig. 9 III-b) and the biker on the bike (Compare Fig. 9 II-d and Fig. 9 III-d). However, a big

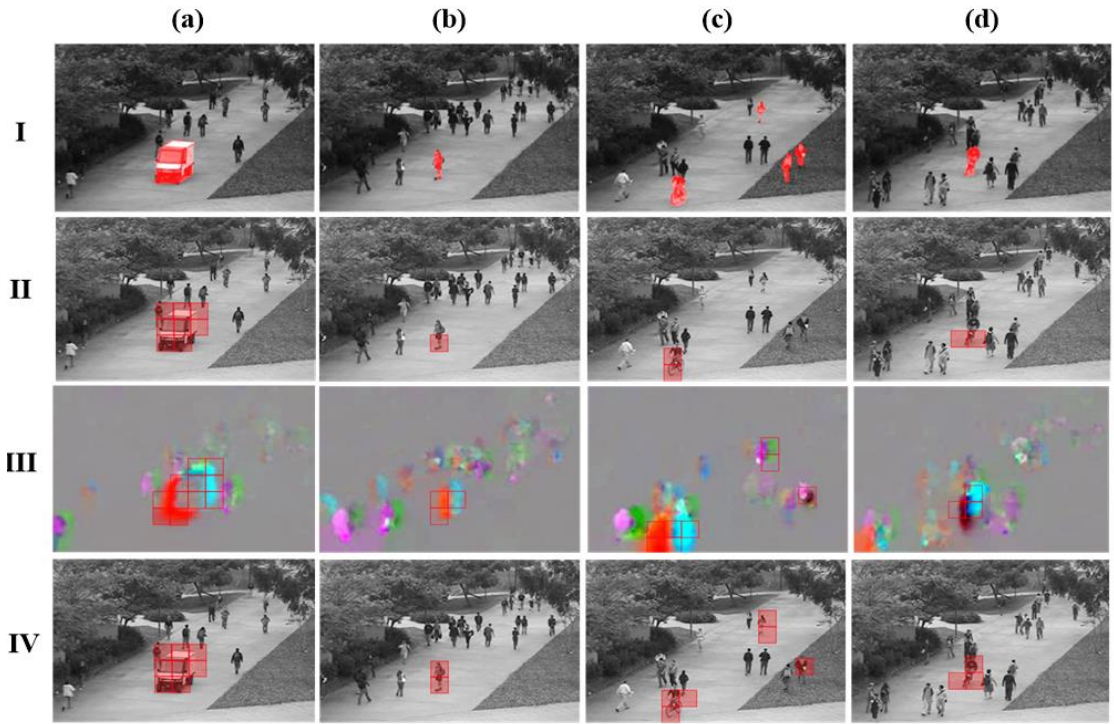

Fig. 9. Examples of the spatial stream and temporal stream detection results on the UCSD Ped1 dataset, in which detected abnormal events are labeled with red masks (I, II, V) or red rectangle (III). (I) Ground-truth. (II) Ours (Appearance) (III) Ours (Motion). (V) Ours (Late fusion)

disadvantage of using just the temporal stream, is that compared to the ground truth it produces some false detections (Fig. 9. III) as a result of complex motion and occlusion. As shown in Fig. 9. IV, combining the spatial and temporal streams (late fusion, as given in (10)) can compromise the misdetection of the spatial stream and false detection of the temporal stream. By combining motion and appearance cues, the detection accuracy can be greatly improved.

## V. CONCLUSION

In this paper, we presented an effective partially supervised deep learning methodology for detecting and locating anomalous events in surveillance videos. . Our approach builds upon a two-stream network framework, which employs RGB frames and dynamic flows, respectively. In the training stage, image patches of normal samples for each stream are extracted as input to train a Gaussian Mixture Fully Convolutional Variational Autoencoder (GMFC-VAE) that learns a Gaussian Mixture Model (GMM). In the testing stage, the conditional probabilities of each component of a Gaussian Mixture of test patches are obtained by employing the GMFC-VAE for each stream. We introduce a sample energy based method for predicting an appearance and motion anomaly score. These two cues are then fused to achieve the final detection results. Both the qualitative and quantitative results on two challenging datasets show that our method outperforms the state-of-the-art methods. In the future, we intend to focus on extending the method by fusing the appearance and temporal streams at an earlier stage in order to provide a simpler network for detecting anomalies.